\documentclass{article}

\usepackage{soul}
\usepackage{color, xcolor}
\definecolor{lightgray}{rgb}{0.7, 0.7, 0.7}

\usepackage[final, dandb, nonatbib]{neurips_2025}
\usepackage[numbers]{natbib}
\usepackage[utf8]{inputenc} 
\usepackage[T1]{fontenc}    
\usepackage{hyperref}       
\usepackage{url}            
\usepackage{booktabs}       
\usepackage{amsfonts}       
\usepackage{nicefrac}       
\usepackage{microtype}      
\usepackage{times}
\usepackage[utf8]{inputenc} 
\usepackage[T1]{fontenc}    
\usepackage{url}            
\usepackage{booktabs}       
\usepackage{amsfonts}       
\usepackage{nicefrac}       
\usepackage{microtype}      
\usepackage{xcolor}         

\usepackage{graphicx}
\usepackage{amsmath,amssymb}
\usepackage{verbatim}
\usepackage{multirow}
\usepackage{marvosym}
\usepackage{makecell}
\usepackage{bm}
\usepackage{wrapfig}
\usepackage{caption}
\usepackage{twemojis}
\usepackage{utfsym}
\usepackage{float}
\usepackage{subfig}
\usepackage{pifont}
\usepackage{enumitem}
\usepackage{amsmath}
\usepackage{geometry}
\usepackage{wrapfig}
\usepackage{colortbl}
\usepackage{etoolbox}

\NewDocumentCommand{\todo}
{ mO{} }{\textcolor{magenta}{\textsuperscript{\textit{TODO}}\textsf{\textbf{\small[#1]}}}}

\title{Impromptu VLA: Open Weights and Open Data for Driving Vision-Language-Action Models}

\author{
  Haohan Chi\textsuperscript{*,1}, 
  Huan-ang Gao\textsuperscript{*,1},
  Ziming Liu\textsuperscript{\textdagger,2}, 
  Jianing Liu\textsuperscript{1}, \\
  Chenyu Liu\textsuperscript{1},
  Jinwei Li\textsuperscript{1}, 
  Kaisen Yang\textsuperscript{1},
  Yangcheng Yu\textsuperscript{1},
  Zeda Wang\textsuperscript{1},
  Wenyi Li\textsuperscript{1}, \\
  Leichen Wang\textsuperscript{2},
  Xingtao Hu\textsuperscript{2},
  Hao Sun\textsuperscript{2},
  Hang Zhao\textsuperscript{3},
  Hao Zhao\textsuperscript{1,\textdagger} \\ \\
  \textsuperscript{1}AIR, Tsinghua University \quad
    \textsuperscript{2}Bosch Research \quad
    \textsuperscript{3}IIIS, Tsinghua University \\
    \textsuperscript{*}Equal contribution \quad \textsuperscript{\textdagger}Corresponding author\\
    Project Page: \url{http://Impromptu-VLA.c7w.tech/}
}

\begin{document}

\maketitle

\begin{figure}[h]
    \centering
    \includegraphics[width=0.9\linewidth]{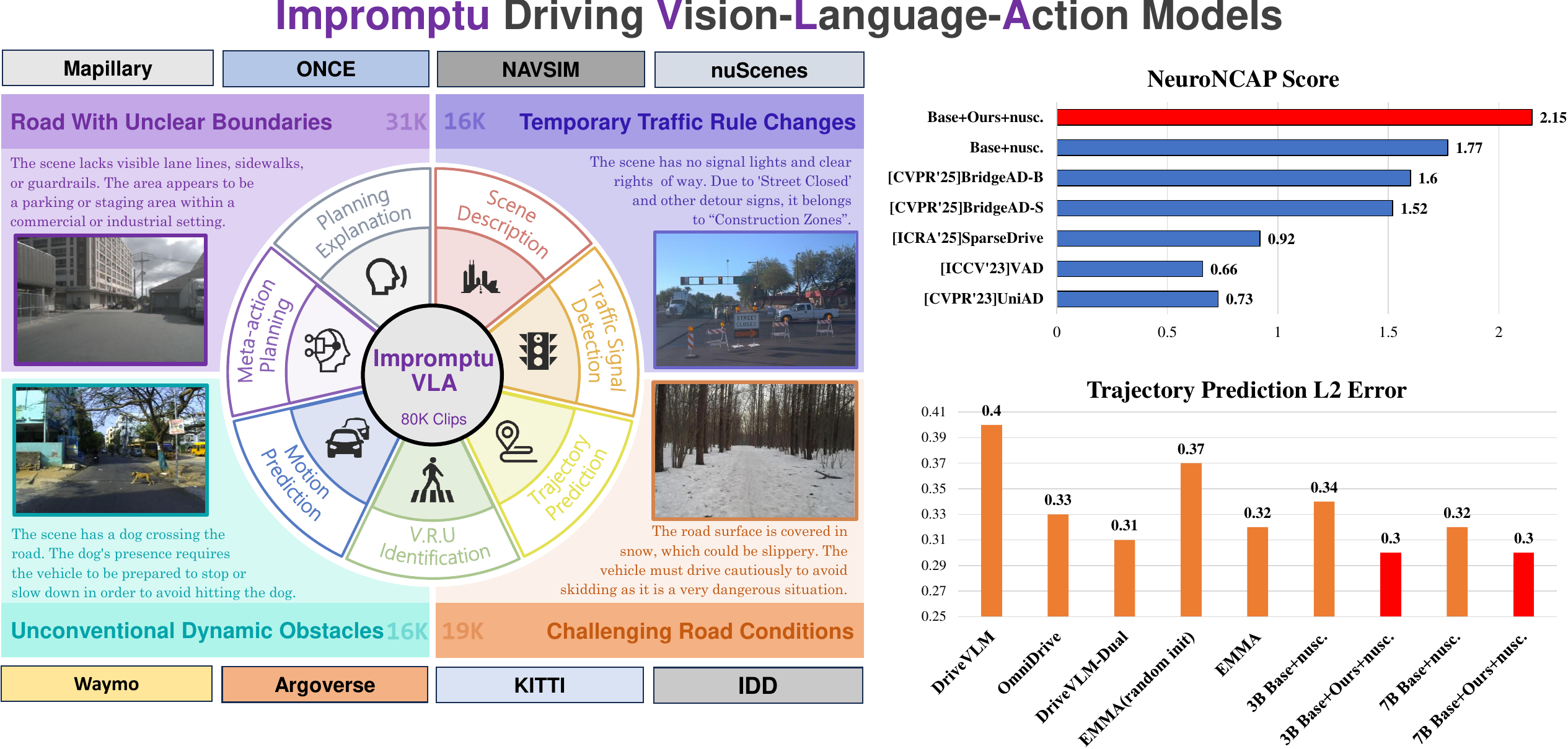}
    \caption{
    \textbf{Visual Abstract of Impromptu VLA.} 
    We construct Impromptu VLA Dataset, which contains over 80K clips curated from 8 open-sourced datasets, focusing on four critical types of unstructured \textit{"corner case"} scenarios that challenge current autonomous driving vehicles. It supports interconnected VLA tasks including scene understanding, prediction, meta planning and trajectory planning. Key experimental results demonstrates that VLA models trained with Impromptu VLA Dataset achieve significant performance improvements in both closed-loop and open-loop metrics.
    }
    \label{fig:abstract}
\end{figure}

\begin{abstract}
Vision-Language-Action (VLA) models for autonomous driving show promise but falter in unstructured \textit{corner case} scenarios, largely due to a scarcity of targeted benchmarks. To address this, we introduce Impromptu VLA. Our core contribution is the \textbf{Impromptu VLA Dataset}: over 80,000 meticulously curated video clips, distilled from over 2M source clips sourced from 8 open-source large-scale datasets. This dataset is built upon our novel taxonomy of four challenging unstructured categories and features rich, planning-oriented question-answering annotations and action trajectories. 
Crucially, experiments demonstrate that VLAs trained with our dataset achieve substantial performance gains on established benchmarks—improving closed-loop NeuroNCAP scores and collision rates, and reaching near state-of-the-art L2 accuracy in open-loop nuScenes trajectory prediction. Furthermore, our Q\&A suite serves as an effective diagnostic, revealing clear VLM improvements in perception, prediction, and planning. 
Our code, data and models are available at \url{https://github.com/ahydchh/Impromptu-VLA}.
\end{abstract}

\section{Introduction}

Autonomous driving has achieved remarkable progress, demonstrating increasing proficiency in navigating the well-structured environments of urban centers and highways where clear lane markings and predictable traffic flows are the norm \cite{uniad, vad, tian2024drivevlm}. However, the ultimate ambition of ubiquitous self-driving compels us to look beyond these well-trodden paths towards the intricate and often unpredictable domain of unstructured roads. These \textit{unstructured} scenarios—encompassing everything from rural tracks and dynamic construction zones to areas with ambiguous signage or those recovering from natural events—represent the next significant frontier. It is here that current autonomous systems often face their sternest tests, and where breakthroughs are essential for realizing the full potential of go-anywhere autonomous capabilities \cite{xie2025drivebench}.

Successfully navigating this frontier is profoundly hindered by a critical scarcity of specialized data. While numerous driving datasets have been foundational to current progress, they predominantly capture common, structured traffic situations \cite{caesar2020nuscenes,nuplan,kitti,Once,neuhold2017mapillary,sun2020scalability,varma2019idd,argoversev2}. This leaves a significant blind spot concerning the sheer diversity and unique challenges posed by unstructured settings, such as ill-defined road boundaries, the appearance of unconventional dynamic obstacles, adherence to makeshift traffic rules, or dealing with treacherous road surfaces. Without large-scale, meticulously annotated datasets that specifically reflect these complex conditions~\cite{wu2023nuprompt, qian2024nuscenes-qa}, the ability to train robust AI drivers and rigorously evaluate their adaptability in such scenarios remains severely constrained.

To address this data void, we introduce the \textbf{Impromptu VLA Dataset}, a new large-scale benchmark specifically curated to propel research in autonomous driving on unstructured roads, as introduced in Figure \ref{fig:abstract}. Distilled from an initial pool of over two million clips from eight diverse public sources~\cite{caesar2020nuscenes,nuplan,kitti,Once,neuhold2017mapillary,sun2020scalability,varma2019idd,argoversev2}, Impromptu VLA comprises approximately $\sim$80,000 meticulously selected and verified clips. These are categorized into four distinct types of challenging unstructured scenarios—roads with unclear boundaries, temporary traffic rule changes, unconventional dynamic obstacles, and challenging road conditions—and are enriched with extensive multi-task annotations and planning trajectories. The dataset was constructed using an advanced pipeline that leverages Vision-Language Models (VLMs) with Chain-of-Thought reasoning~\cite{liu2024llava, bai2023qwenvl, chen2024internvl} for nuanced understanding, followed by comprehensive human verification to ensure high-quality, reliable labels.

Our comprehensive experimental evaluations rigorously validate the efficacy of the Impromptu VLA Dataset. We demonstrate that VLMs fine-tuned on our dataset exhibit substantially improved performance on established autonomous driving benchmarks. For instance, in challenging closed-loop NeuroNCAP \cite{ljungbergh2024neuroncap} simulations (Table \ref{table:NeuroNCAP}), our 3B model enhanced with Impromptu VLA saw its average NeuroNCAP score increase significantly to \textbf{2.15}/5.00 from 1.77/5.00 achieved by the baseline, while its average collision rate was critically reduced from 72.5\% down to \textbf{65.5\%}. 
In open-loop nuScenes \cite{caesar2020nuscenes} evaluations for trajectory prediction, pre-training with our dataset also markedly reduced L2 errors; our 3B model fine-tuned with Impromptu VLA achieved an average L2 error of 0.30m, bringing its performance nearly on par with leading specialized methods like EMMA+ \cite{hwang2024emma} (0.29m), despite the latter often benefiting from substantially larger proprietary training datasets~\cite{zhou2024elm, driveadapter}. 
Furthermore, evaluations on our dataset's own diverse Q\&A validation suite reveal significant and quantifiable gains in specific VLM capabilities related to perception, prediction, and planning within these demanding unstructured contexts.

Our primary contributions are summarized as follows:

\begin{itemize}
    \item The Impromptu VLA Dataset: A publicly available, large-scale, and richly annotated resource meticulously focused on diverse and challenging unstructured driving scenarios, designed to fill a critical gap in existing data resources.
    \item A systematic taxonomy for unstructured road conditions and a scalable, VLM-centric data curation pipeline for their identification, categorization, and comprehensive annotation with multi-task Q\&A suitable for training advanced VLMs.
    \item Extensive experimental evidence demonstrating that training with the Impromptu VLA Dataset significantly boosts results on standard driving benchmarks, and serves as an effective diagnostic tool for assessing and improving VLM capabilities in unstructured environments.
\end{itemize}
\vspace{-20pt}

\section{Impromptu VLA Dataset: Learning to Drive on Unstructured Roads} 

\begin{figure}[t]
    \centering
    \includegraphics[width=1\linewidth]{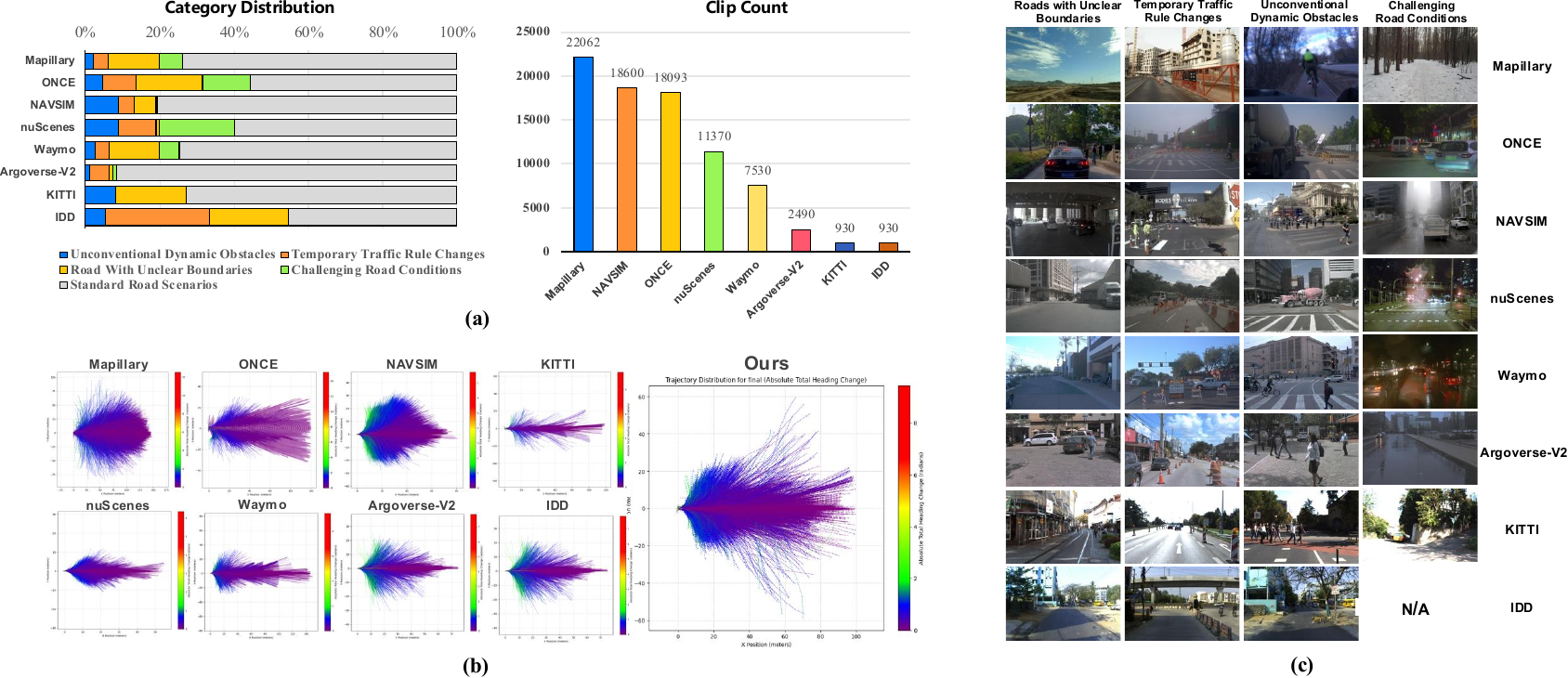}
    
    \caption{Characteristics comparison of different driving scene datasets. Figure (a) illustrates the distribution of scene categories across various datasets and the number of video clips contained in each dataset, providing a direct view of the emphasis on different scene types and the data scale of each dataset. Figure (b) compares the trajectory distribution in the original with the trajectory distribution in our constructed dataset, explaining the trajectory diversity of our dataset. Figure (c) shows examples of different scene categories from 8 source datasets. Notably, the IDD dataset lacks data for the "Challenging Road Conditions" category.}
    \label{fig:statistics}
\end{figure}

\begin{table}[t]
\centering
\footnotesize
\resizebox{1.0\linewidth}{!}{%
    \begin{tabular}{l c c c c c c c c c}
    \toprule
    Attribute & Mapillary & ONCE & NAVSIM & nuScenes & Waymo & Argoverse-V2 & KITTI & IDD & Sum. \\
    \midrule
    Camera Views & 1 & 7 & 8 & 6 & 5 & 7 & 1 & 1 & - \\
    Resolutions & Variable & 1920 x 1020 & 1920 x 1080 & 1600 x 900 & 640 x 360 & 1550 x 2048 & 1242 x 375 & 1920 x 1080 & - \\ 
    FPS & 2Hz* & 2Hz & 2Hz & 2Hz & 10Hz & 20Hz & 10Hz & 15Hz & \\
    Raw Clip Count & 1000k & 800k & 90k & 40k & 40k & 320k & 20k & 7k & 2000k \\
    Labeled Clip Count & 22062 & 18093 & 18600 & 11370 & 7530 & 2490 & 930 & 930 & 80k\\
    Raw Data Size & 60GB & 2TB & 450GB & 1.5TB & 5TB & 1TB & 180GB & 18GB & 10TB\\
    Labeled Data Size & 1GB & 5GB & 12GB & 10GB & 7GB & 7GB & 1GB & 0.5GB & 43.5GB \\ 
    \bottomrule
\end{tabular}
}
\caption{
    \textbf{Dataset Information.} Summary of key attributes for the datasets used in this study. Note that the Mapillary dataset exhibits variable resolutions. For the Mapillary dataset, the frequency is assumed to be 2Hz as specific FPS information is not explicitly provided.
    }
\label{table:dataset_info}
\end{table}

\subsection{Overview}
\begin{figure}[t]
    \centering
    \includegraphics[width=1\linewidth]{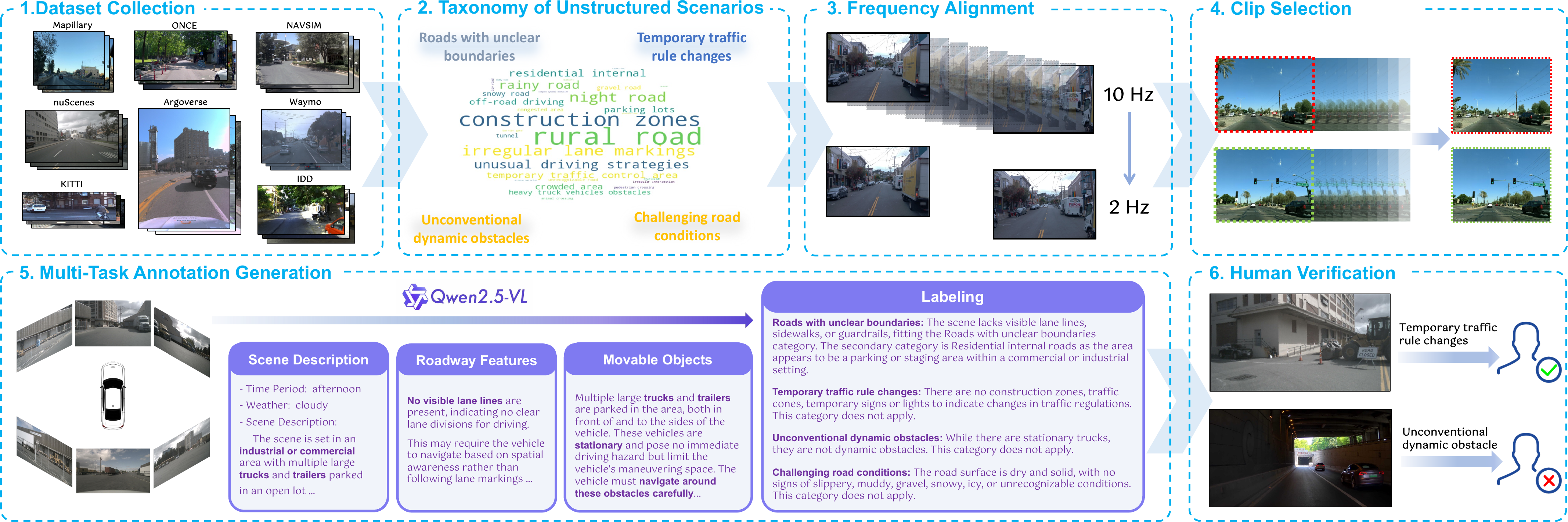}
    \caption{
    \textbf{Data Processing and Annotation Pipeline for the Impromptu VLA Dataset.} The diagram outlines the sequential process for creating our dataset, starting from raw data collection and scenario taxonomy definition (Sec.~\ref{sec:Taxonomy}, through frequency alignment and keyclip selection, to multi-task annotation generation via Qwen2.5-VL (including scene description, object/feature analysis, and labeling), and concluding with rigorous human verification (Sec.~\ref{sec:task}).
    }
    \label{fig:label-pipe}
\end{figure}

The research community currently lacks sufficient large-scale, diverse, and meticulously annotated datasets specifically focused on unstructured scenarios. 
To address this critical gap, we introduce the \textbf{Impromptu VLA Dataset}, a dataset curated to foster advancements in autonomous driving on unstructured roads. Sourced from an initial aggregation of over 2 million clips (occupying over 10T of storage) from eight prominent public datasets \cite{caesar2020nuscenes,nuplan,kitti,Once,neuhold2017mapillary,sun2020scalability,varma2019idd,argoversev2}, the Impromptu VLA Dataset has been distilled into a highly concentrated collection of $\sim$80,000 clips after our selection mechanism, which is illustrated in Figure \ref{fig:label-pipe}. The resulting dataset specifically captures a diverse array of challenging scenarios, including roads with unclear boundaries, the presence of unconventional dynamic obstacles, and segments with temporary or non-standard traffic rules (see Table \ref{table:dataset_info} for detailed statistics).

\subsection{Defining a Taxonomy for Unstructured Driving Scenarios}
\label{sec:Taxonomy}
A primary objective in creating the Impromptu VLA Dataset was to move beyond a monolithic and vague view of \textit{unstructuredness} and establish a more granular understanding of the specific challenges these environments present. To achieve this, and to focus the dataset on scenarios that genuinely test the limits of current autonomous driving systems, our preliminary effort undertook a data-driven process to define a concise yet comprehensive taxonomy of unstructured road scenarios. 

Our methodology for defining these categories began with an extensive, unbiased exploration of the collected data. We first created a representative subset by sampling approximately 10\% of the clips at regular intervals from the aggregated and standardized multi-source dataset. This subset was then subjected to an open-ended descriptive analysis using the capabilities of a powerful Vision-Language Model, Qwen2.5-VL 72B \cite{bai2025qwen2}. Instead of querying the model to answer questions in a predefined label protocol, we leveraged the VLM's advanced image understanding capabilities to prompt it to generate detailed textual descriptions for each scene, as shown in the Appendix. 

The subsequent phase involved a multi-stage, highly automated process to distill these descriptions into meaningful categories of unstructured challenges. First, to \textbf{programmatically identify and filter out conventional driving scenarios}, we employed another VLM-based classification step. Each initial, rich scene description generated by Qwen2.5-VL was evaluated using a carefully designed prompt, which instructed the VLM to act as a scenario categorizer to judge if the caption belongs to unconventional cases. 
To ensure the reliability and effectiveness of this VLM-based filtering prompt, we conducted an iterative refinement process of prompt. This process is tested on a validation subset of $\sim$1000 scene descriptions, which were also manually and independently labeled as \textit{'Conventional'} or \textit{'Unconventional'} by two human annotators. The VLM's classifications were compared against human consensus, and the prompt was iteratively adjusted until achieving a high degree of agreement.

For the selected unconventional scenarios from the full set, we conduct semantic-level analysis to identify recurring patterns and group semantically similar unstructured scenarios. This clustering allowed for the bottom-up emergence of potential subcategories, such as those involving \textit{"unclear road edges," "temporary road work," "animals on road," or "poor visibility due to snow."} Through iterative refinement, consolidation of these machine-generated clusters, and abstraction based on the primary source of driving complexity identified in these groups, we converged on the four salient high-level categories detailed below.

\textit{1. Roads with unclear boundaries}: Scenarios where the traversable path is ambiguous or undefined, such as rural dirt tracks, off-road trails, or roads with faded/absent markings. These severely challenge perception tasks like lane detection and drivable area segmentation.

\textit{2. Temporary traffic rule changes}: Dynamic situations where standard traffic rules are temporarily altered by construction zones, human traffic controllers, or temporary signage, requiring autonomous vehicles to adapt to unusual instructions and road layouts.

\textit{3. Unconventional dynamic obstacles}: Features dynamic actors or obstacles uncommon in typical urban driving that demand specialized interaction strategies. Examples include large or erratically moving vehicles, vulnerable road users in unexpected locations, or animal encounters, all posing sudden hazards.

\textit{4. Challenging road conditions}: Encompasses scenarios where adverse road surfaces (e.g., potholes, mud, snow, ice) or environmental conditions (e.g., fog, heavy rain, low-light, glare) severely impair visibility or affect vehicle dynamics, complicating hazard perception and safe navigation.


\subsection{Data Processing and Annotation}
\label{sec:task}

Following the definition of our unstructured scenario taxonomy (Section \ref{sec:Taxonomy}), the curated data underwent several processing and annotation stages as depicted in Figure \ref{fig:label-pipe}. 

\textbf{Keyclip Selection and Stability Filtering.}
All collected sequences were first standardized to a uniform temporal rate of 2 Hz, addressing inconsistencies from diverse sources (Table \ref{table:dataset_info}).
We aligned the clip configuration with NAVSIM \cite{dauner2024navsim}, keeping 1.5 seconds from the past and 5 seconds for the future, and selected that central keyclip from each pack for annotation.
To minimize false positives from transient keyclip-level predictions, we employed a temporal stability packing mechanism. Specifically, adjacent clips were packed into (up to if possible) 15-second "local-filter pack". Scene characteristic of a clip (preliminarily identified at keyclip-level) was only considered stable and propagated to subsequent annotation stages if it persisted for a minimum number of clips within this pack (e.g., more than a single occurrence). It is important to note that these "local-filter pack" were solely for this stability check and selection process; the final dataset primarily consists of individually annotated keyclips.

\textbf{Scene Classification and Structured Information Extraction via CoT prompting.}
Selected keyclips were classified using Qwen2.5-VL 72B \cite{bai2025qwen2} with Chain-of-Thought (CoT) prompting \cite{wei2022chain} to extract rich structured information beyond simple captions. This hierarchical reasoning process analyzed overall scene context (R1: description), static roadway features (R2), movable objects (R3), and culminated in a justified final assignment (R4) to one of our four unstructured scene categories (Section 3.2). The structured CoT output provided not only the scene category but also a wealth of contextual information for subsequent task annotation.

\textbf{Multi-Task Annotation Generation.}
Leveraging the scene category and the structured information extracted during the CoT process, we further enriched each keyclip with a diverse set of task-specific annotations, drawing inspiration from comprehensive annotation frameworks like Senna \cite{jiang2024senna}. 
This multi-task annotation was achieved through a combination of rule-based and LLM-based methods. Specifically, we generated the following annotations for each selected keyclip.
\textit{1. Scene Description}: Comprehensive descriptions capturing the overall environmental context, time, weather, and traffic conditions were produced through targeted queries to VLM.
\textit{2. Traffic Signal Detection}: The presence state and type of active traffic signals were identified via further VLM queries.
\textit{3. Vulnerable Road User (VRU) Identification}: Information on VRUs, including their presence, type (e.g., pedestrian, cyclist), and distances from the ego vehicle, was derived from ground truth data.
\textit{4. Motion Intention Prediction}: To capture dynamic aspects, predicted motion intentions for key actors in the scene were generated by VLM.
\textit{5. Meta-action Planning}: High-level plans (e.g., accelerate-left, keep-straight) for the ego-vehicle were formulated, typically through VLM prompting conditioned on the scene context.
\textit{6. Planning Explanation}: Textual explanations, rationalizing potential or actual ego-vehicle maneuvers in response to the scene, were generated by the VLM.
\textit{7. End-to-End Trajectory Prediction}: Data to support this task was curated by structuring past vehicle states and corresponding future target trajectories in the ground truth.

\textbf{Comprehensive Human Verification.}
All generated annotations—both the primary unstructured scene category and the subsequent multi-task labels—were subjected to a meticulous human verification process. Annotators reviewed each keyclip and its associated labels, providing a binary judgment (accept/reject) or performing minor corrective edits if necessary. This ensured high fidelity across the entire dataset. To quantitatively assess the VLM's scene classification performance for our defined unstructured categories prior to extensive human review, we evaluated it on a subset of 200 images sampled at intervals from the nuScenes dataset. Comparing VLM classifications against expert manual labels yielded strong F1 scores for several categories: 0.90 for 'Temporary Traffic Rule Changes', 0.81 for 'Unconventional Dynamic Obstacles', and 0.91 for 'Challenging Road Conditions'. The 'Road With Unclear Boundaries' category was found to be too rare within this specific nuScenes subset for a meaningful F1 score calculation. These validation results provide confidence in the VLM-based stages of our annotation pipeline. 

\subsection{Dataset Statistics}
The final Impromptu VLA Dataset comprises a substantial collection of annotated clips specifically curated for their unstructured road characteristics. Figure \ref{fig:statistics} illustrates the total number of these clips derived from each source dataset and presents the overall distribution of these clips across the four defined unstructured scenario categories introduced in Sec.~\ref{sec:Taxonomy}. The coverage of trajectory distribution is also reported in Figure \ref{fig:statistics}.

To maximize the utility of this dataset for training and evaluating perception and planning models, the rich multi-task annotations generated for each clip (as detailed in Sec.~\ref{sec:task}) are structured as planning-oriented Question-Answering (Q\&A) pairs. 
This format, inspired by frameworks like DriveVLM \cite{tian2024drivevlm} or EMMA \cite{hwang2024emma}, directly associates visual inputs, text outputs and action trajectory predictions within sequence space of LLMs.  
For standardized evaluation, the entire dataset of curated clips, across all four unstructured categories, is partitioned into training and validation sets using an 80:20 split. This stratification is performed within each category to ensure that the validation set maintains a representative distribution of all defined unstructured road challenges.

\section{Experiments}
This section empirically validates our Impromptu VLA Dataset by investigating its impact on advancing autonomous driving models. We seek to answer:

(1) Does training with our dataset improve vision-language model (VLM) performance on existing benchmarks, both closed-loop and open-loop?

(2) In which specific aspects does the Impromptu VLA Dataset enhance VLM performance - perception, prediction, or planning? How effectively does our validation set, with its detailed planning-oriented Q\&A, serve as a diagnostic benchmark for pinpointing these contributions and evaluating model capabilities in these distinct tasks?


\begin{table*}[t]
\centering
\footnotesize
\resizebox{0.95\linewidth}{!}{%
\begin{tabular}{ll@{\hspace{1.5em}}cccc@{\hspace{1.5em}}cccc} 
\toprule
\multirow{2}{*}{\textbf{Source}} & \multirow{2}{*}{\textbf{Method}} & \multicolumn{4}{c}{\textbf{NeuroNCAP Score $\uparrow$}} & \multicolumn{4}{c}{\textbf{Collision rate (\%) $\downarrow$}} \\
\cmidrule(lr){3-6} \cmidrule(lr){7-10} 
& & \textbf{Avg.} & \textbf{Stat.} & \textbf{Frontal} & \textbf{Side} & \textbf{Avg.} & \textbf{Stat.} & \textbf{Frontal} & \textbf{Side} \\ 
\midrule
CVPR 2023 & UniAD\textsuperscript{2}                         & 0.73          & 0.84   & 0.10    & 1.26   & 88.6          & 87.8   & 98.4   & 79.6   \\
ICCV 2023 & VAD\textsuperscript{2}                           & 0.66          & 0.47   & 0.04    & 1.45   & 92.5          & 96.2   & 99.6   & 81.6   \\

ICRA 2025 & SparseDrive\textsuperscript{1}                   & 0.92          & -      & -       & -      & 93.9          & -      & -      & -      \\
CVPR 2025 & BridgeAD-S\textsuperscript{1}                    & 1.52          & -      & -       & -      & 76.2          & -      & -      & -      \\
CVPR 2025 & BridgeAD-B\textsuperscript{1}                    & 1.60          & -      & -       & -      & 72.6          & -      & -      & -      \\
\cellcolor{gray!20!white}- & \cellcolor{gray!20!white}Base+nuScenes                 & \cellcolor{gray!20!white}\underline{1.77} & \cellcolor{gray!20!white}\textbf{1.80} & \cellcolor{gray!20!white}\underline{1.67} & \cellcolor{gray!20!white}\underline{1.75} & \cellcolor{gray!20!white}\underline{72.5} & \cellcolor{gray!20!white}\textbf{68.0} & \cellcolor{gray!20!white}\underline{73.0} & \cellcolor{gray!20!white}\underline{71.5} \\
\cellcolor{gray!20!white}- & \cellcolor{gray!20!white}\textbf{Base+Impromptu+nuScenes} & \cellcolor{gray!20!white}\textbf{2.15} & \cellcolor{gray!20!white}\underline{1.77} & \cellcolor{gray!20!white}\textbf{2.31} & \cellcolor{gray!20!white}\textbf{2.10} & \cellcolor{gray!20!white}\textbf{65.5} & \cellcolor{gray!20!white}\underline{70.0} & \cellcolor{gray!20!white}\textbf{59.0} & \cellcolor{gray!20!white}\textbf{65.0} \\
\bottomrule
\end{tabular}
}
\caption{Results on NeuroNCAP. (where \textsuperscript{1} indicates sourced from \cite{zhang2025bridging} and \textsuperscript{2} indicates sourced from \cite{ljungberghneural}) Best scores in each category (without/with post-processing) are in \textbf{bold}, second best are \underline{underlined}. The improvements in both the overall NeuroNCAP score and, crucially, the reduction in collision rates suggest that our dataset helps the model develop a more nuanced understanding of complex road interactions, leading to more robust and safer driving policies.}
\label{table:NeuroNCAP}
\end{table*}

\subsection{Pushing Forward Boundaries of Existing End-to-end Autonomous Driving Benchmarks}

\textbf{Closed-loop evaluation.} 
We choose NeuroNCAP \cite{ljungbergh2024neuroncap}, a comprehensive closed-loop evaluation framework that leverages the nuScenes dataset to simulate a wide array of challenging real-world driving scenarios, allowing for the assessment of an autonomous vehicle's planning and control systems in terms of safety and efficiency under diverse conditions. The NeuroNCAP evaluation quantifies performance primarily through collision rates and the NeuroNCAP score (NNS). The NNS is computed, in the spirit of a 5-star rating system, as follows: a score of $5.0$ is achieved if no collision occurs; otherwise, the score is $4.0 \cdot \max(0, 1 - v_i / v_r)$, where $v_i$ is the actual impact speed (magnitude of relative velocity between the ego-vehicle and the colliding actor) and $v_r$ is the reference impact speed that would occur if no evasive action were performed. This means that if a collision is not avoided, the score linearly decreases from a potential 4 points towards 0 as the impact speed $v_i$ approaches or exceeds the reference speed $v_r$. Collision rates, on the other hand, directly track the percentage of scenarios resulting in a collision. These two metrics are categorized by interaction types (e.g., frontal, side).


Our method involves a comparative study of two distinct training pipelines. The base model here is Qwen2.5VL 3B \cite{bai2025qwen2}. The first pipeline, which we term \textbf{"Base+Impromptu+nuScenes"} in Table \ref{table:NeuroNCAP}, involves initially fine-tuning the base VLM on the training split of our Impromptu VLA dataset, and subsequently further fine-tuning this adapted model on the nuScenes training set. 
The second pipeline, \textbf{"Base+nuScenes"}, directly fine-tunes the base VLM on the nuScenes training set without any exposure to the Impromptu VLA. Both models are then evaluated on the NeuroNCAP benchmark.

\begin{wrapfigure}[42]{R}{0.6\textwidth} 
    \centering 
    \footnotesize
    \captionsetup{type=table}
    \vspace{-15pt}
    \resizebox{\linewidth}{!}{
        \begin{tabular}{lccccc} 
            \toprule
            \multirow{2}{*}{\textbf{Method}} & \multicolumn{4}{c}{\textbf{L2 Error (m) $\downarrow$}} \\
            \cmidrule(lr){2-5}
            & 1s & 2s & 3s & \cellcolor{blue!10!white}\textbf{Avg.} \\ 
            \midrule
            \multicolumn{5}{l}{\textit{Closed-source API-only Models}} \\
            \midrule
            GPT-4o\textsuperscript{1} \cite{hurst2024gpt} & \textbf{0.28} & \textbf{0.93} & \textbf{2.02} & \cellcolor{blue!10!white}\textbf{1.07} \\ 
            Claude-3.5-Sonnet\textsuperscript{1} & \underline{0.29} & 0.98 & 2.12 & \cellcolor{blue!10!white}1.13 \\ 
            Claude-3.7-Sonnet\textsuperscript{1} & \textbf{0.28} & \underline{0.94} & \underline{2.04} & \cellcolor{blue!10!white}\underline{1.09} \\ 
            Gemini-2.0-Flash\textsuperscript{1} & 0.31 & 1.08 & 2.36 & \cellcolor{blue!10!white}1.25 \\ 
            Gemini-2.5-Pro\textsuperscript{1} & 0.37 & 1.35 & 2.96 & \cellcolor{blue!10!white}1.56 \\ 
            \midrule
            \multicolumn{5}{l}{\textit{Open-source Generalist VLMs}} \\
            \midrule
            LLaVA-1.6-Mistral-7B\textsuperscript{2} & 1.49 & 3.38 & 4.09 & \cellcolor{blue!10!white}2.98 \\ 
            Llama-3.2-11B-Vision-Instruct\textsuperscript{2} & 1.54 & 3.31 & 3.91 & \cellcolor{blue!10!white}2.92 \\ 
            Qwen2-VL-7B-Instruct\textsuperscript{2} \cite{Qwen2-VL-7B-Instruct} & 1.45 & 3.21 & 3.76 & \cellcolor{blue!10!white}2.81 \\ 
            DeepSeek-VL2-16B\textsuperscript{1} \cite{DeepSeek-VL2}& 0.66 & 1.68 & 2.92 & \cellcolor{blue!10!white}1.75 \\ 
            DeepSeek-VL2-28B\textsuperscript{1} \cite{DeepSeek-VL2}& \textbf{0.37} & \underline{1.35} & 2.96 & \cellcolor{blue!10!white}1.56 \\ 
            LLaMA-3.2-11B-Vision-Instruct\textsuperscript{1}& 0.52 & 1.42 & \underline{2.68} & \cellcolor{blue!10!white}\underline{1.54} \\ 
            LLaMA-3.2-90B-Vision-Instruct\textsuperscript{1}& 0.66 & 1.71 & 3.01 & \cellcolor{blue!10!white}1.79 \\ 
            Qwen-2.5-VL-7B-Instruct\textsuperscript{1} \cite{Qwen2.5-VL-7B-Instruct}& \underline{0.46} & \textbf{1.33} & \textbf{2.55} & \cellcolor{blue!10!white}\textbf{1.45} \\ 
            \midrule
            \multicolumn{5}{l}{\textit{Training-based Driving Specialists (Existing Methods)}} \\
            \midrule
            UniAD\textsuperscript{3} \cite{uniad}& 0.42 & 0.64 & 0.91 & \cellcolor{blue!10!white}0.66 \\ 
            VAD\textsuperscript{3} \cite{vad}& 0.17 & 0.34 & 0.60 & \cellcolor{blue!10!white}0.37 \\ 
            BEV-Planner\textsuperscript{3} \cite{li2024ego}& \underline{0.16} & \textbf{0.32} & \textbf{0.57} & \cellcolor{blue!10!white}\textbf{0.35} \\ 
            Ego-MLP\textsuperscript{3}* \cite{li2024ego}& \textbf{0.15} & \textbf{0.32} & \underline{0.59} & \cellcolor{blue!10!white}\textbf{0.35} \\ 
            \midrule
            \multicolumn{5}{l}{\textit{Ours and Key Competitors (Specialized Driving Models)}} \\
            \midrule
            DriveVLM\textsuperscript{3} \cite{tian2024drivevlm}& 0.18 & 0.34 & 0.68 & \cellcolor{blue!10!white}0.40 \\ 
            OmniDrive\textsuperscript{3} \cite{wang2024omnidrive}& \underline{0.14} & 0.29 & 0.55 & \cellcolor{blue!10!white}0.33 \\ 
            DriveVLM-Dual\textsuperscript{3} \cite{tian2024drivevlm}& 0.15 & 0.29 & \textbf{0.48} & \cellcolor{blue!10!white}0.31 \\ 
            EMMA (random init) \cite{hwang2024emma}\textsuperscript{3} & 0.15 & 0.33 & 0.63 & \cellcolor{blue!10!white}0.37 \\ 
            EMMA \cite{hwang2024emma}\textsuperscript{3} & \underline{0.14} & 0.29 & 0.54 & \cellcolor{blue!10!white}0.32 \\ 
           \textcolor{lightgray}{EMMA+}\textsuperscript{3} \cite{hwang2024emma}& \textcolor{lightgray}{0.13} & \textcolor{lightgray}{0.27} & \textcolor{lightgray}{0.48} & \cellcolor{blue!10!white}\textcolor{lightgray}{0.29} \\ 
            3B Base+nuScenes & 0.14 & 0.30 & 0.58 & \cellcolor{blue!10!white}0.34 \\ 
            3B Base+Impromptu+nuScenes & \textbf{0.13} & \textbf{0.27} & \underline{0.52} & \cellcolor{blue!10!white}\textbf{0.30} \\ 
            7B Base+nuScenes & \textbf{0.13} & \underline{0.28} & 0.55 & \cellcolor{blue!10!white}\underline{0.32}\\ 
            7B Base+Impromptu+nuScenes & \textbf{0.13} & \textbf{0.27} & 0.53 & \cellcolor{blue!10!white}\textbf{0.30} \\ 
            \bottomrule
        \end{tabular}%
    } 
    \caption{Open-loop trajectory prediction L2 errors (m) on the nuScenes dataset. (where \textsuperscript{1} indicates sourced from \cite{qiao2025lightemma}, \textsuperscript{2} indicates sourced from \cite{xing2025openemma} and \textsuperscript{3} indicates sourced from \cite{hwang2024emma}). Best results within each category are in \textbf{bold}, second best are \underline{underlined}.}
    \label{table:finetune-L2} 
\end{wrapfigure}

\textbf{Open-loop Evaluation.}
In addition to closed-loop simulations, we conduct open-loop evaluations to specifically assess the trajectory prediction accuracy of VLMs when benefiting from our Impromptu VLA. For this, we also utilize the nuScenes dataset \cite{caesar2020nuscenes}, focusing on the end-to-end trajectory prediction task. Performance is primarily measured by the L2 distance (in meters) between the predicted and ground truth trajectories at future time horizons of 1s, 2s, and 3s, along with the average L2 error.
The experimental methodology mirrors the comparative approach used in the closed-loop tests. We compare two main training strategies for the Qwen2.5VL 3B and 7B VLMs: (1) \textbf{"Base+nuScenes"}, where base VLM is directly fine-tuned on the nuScenes dataset, and (2) \textbf{"Base+Impromptu+nuScenes"}, where base VLM is first fine-tuned on our Impromptu VLA, and this adapted model is then further fine-tuned on nuScenes.
This comparison aims to isolate the benefits conferred by pre-training on our dataset for the task of trajectory prediction in diverse scenarios. The results, contextualized with several state-of-the-art methods, are detailed in Table \ref{table:finetune-L2}.

As demonstrated in Table \ref{table:finetune-L2}, the open-loop trajectory prediction results on the nuScenes benchmark reveal a marked improvement when models are pre-trained on our Impromptu VLA Dataset.
The gains in trajectory accuracy are consistently observed across the 1s, 2s, and 3s prediction horizons. 
Impressively, this level of enhancement brings the performance of our adapted 3B/7B models, into a competitive range with leading methods such as EMMA+ \cite{hwang2024emma} (average L2 of 0.29m), despite EMMA+ benefiting from training on substantially larger internal datasets with millions of scenarios, introduced by Waymo. This underscores the efficacy of the Impromptu VLA Dataset (80K clips) in significantly boosting trajectory prediction capabilities.

\subsection{Diagnostic Evaluation of VLM Capabilities on Impromptu VLA}
To answer the second question—investigating which specific aspects of autonomous driving (perception, prediction, or planning) are enhanced by the Impromptu VLA Dataset and how well our validation set serves as a diagnostic benchmark—we conducted a series of evaluations using its planning-oriented Q\&A tasks. This involved comparing the performance of base Vision-Language Models (VLMs) against versions fine-tuned on our dataset in a task-oriented manner.

\begin{table}[h]
\centering
\footnotesize
\resizebox{0.8\linewidth}{!}{
\begin{tabular}{l@{\hspace{1em}}cccc@{\hspace{1em}}ccccc} 
\toprule
\multirow{2}{*}{\textbf{Method}} & \multicolumn{4}{c}{\textbf{Q\&A Accuracy $\uparrow$}} & \multicolumn{5}{c}{\textbf{Traj. Pred. L2 Error (m) $\downarrow$}} \\ 
\cmidrule(lr){2-5} \cmidrule(lr){6-10} 
& \textbf{V.R.U.} & \textbf{T. Light} & \textbf{Dyn. Obj.} & \textbf{M.P.} & \textbf{1s} & \textbf{2s} & \textbf{3s} & \textbf{4s} & \textbf{Avg.} \\ 
\midrule
3B Base & 0.87 & 0.95 & 0.20 & 0.56 & 3.39 & 5.31 & 7.70 & 10.08 & 6.62 \\ 
3B Base+Impromptu & \textbf{0.91} & \textbf{0.96} & \textbf{0.92} & \textbf{0.84} & \textbf{0.16} & \textbf{0.43} & \textbf{0.82} & \textbf{1.34} & \textbf{0.69} \\ 
\midrule
7B Base & 0.86 & 0.92 & 0.22 & 0.55 & 2.99 & 4.80 & 6.64 & 8.52 & 5.74 \\ 
7B Base+Impromptu & \textbf{0.91} & \textbf{0.97} & \textbf{0.92} & \textbf{0.83} & \textbf{0.10} & \textbf{0.31} & \textbf{0.65} & \textbf{1.11} & \textbf{0.54} \\ 
\bottomrule
\end{tabular}
} 
\caption{
    \textbf{Quantitative Evaluation on the Impromptu VLA validation set.} Performance comparison on various Q\&A tasks within our validation set. The table shows metrics for 3B and 7B Qwen2.5-VL models, comparing a Base version against one fine-tuned on Impromptu VLA ('Ours'). Accuracy $\uparrow$ is reported for perception (V.R.U., T. Light), prediction (Dyn. Obj.), meta-planning (M.P.) and Planning (L2). Best results are in \textbf{bold}.
}
\label{table:QA_res_swapped} 
\end{table}

The quantitative evaluation on the Impromptu VLA validation set, summarized in Table \ref{table:QA_res_swapped}, clearly demonstrates that fine-tuning on our dataset can transforms to all crucial aspects of autonomous driving, including perception, prediction, reasoning for planning, and the planned trajectory.

\begin{figure}
    \centering
    \includegraphics[width=1.0\linewidth]{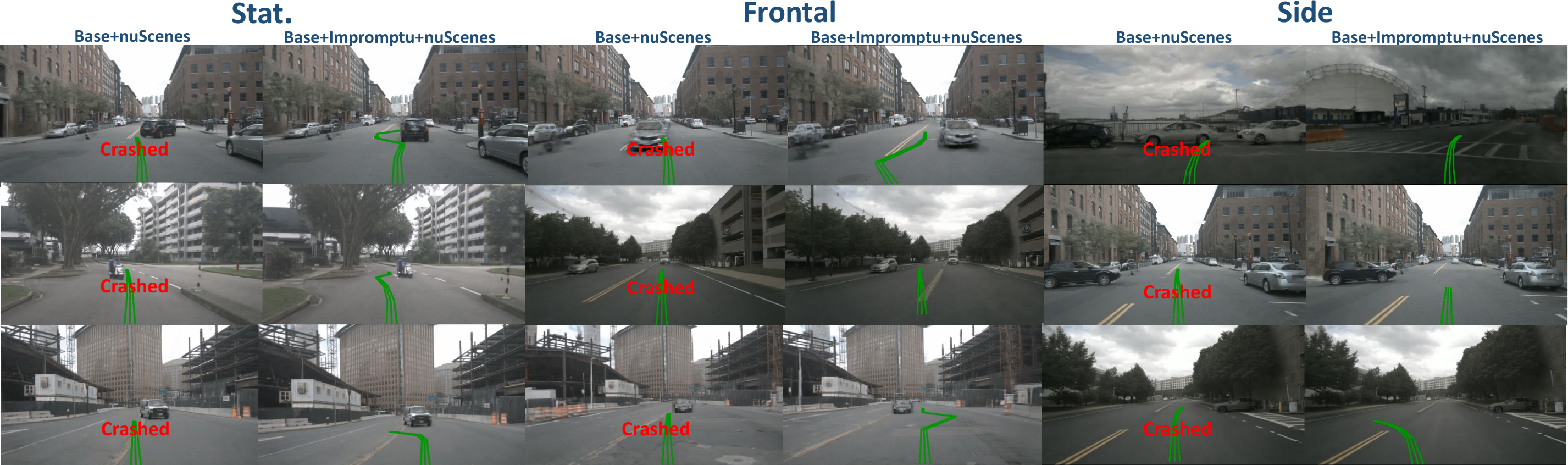}
    \caption{
    \textbf{NeuroNCAP performance in challenging scenarios. } This figure compares the driving behavior of the two models in three representative challenging scenarios: static, frontal, and side. For each scenario, the left column shows the behavior of the base model, which is fine-tuned on nuScenes. The right column shows the performance of the model trained on a subset of our proposed dataset and then fine-tuned on nuScenes. Compared to the base model, the model using our data can better avoid vehicles by turning, slowing down, etc.}
    \label{fig:finetune-traj}
\end{figure}

\section{Related Works}

\textbf{When Vision Language Models Meet Autonomous Driving.}
Vision Language Models (VLMs) extend Large Language Models (LLMs) with visual understanding capabilities~\cite{liu2024llava, bai2023qwenvl, chen2024internvl, wang2023cogvlm, achiam2023gpt4,ding2024hint,jin2023adapt,jin2024tod3cap,xu2024vlm,zhou2024hints,insightdrive2025,hu2022stp3,chen2024vadv2,zheng2024genad,weng2024paradrive,zhang2024sparsead,li2024latentworld, wu2023mars,zheng2023steps,jiang2024p,zheng2024monoocc,yan2023int2,li2024uniscene} enabling multimodal reasoning. These models have recently been introduced into autonomous driving, either to complement traditional end-to-end frameworks~\cite{codevilla2019exploring, uniad, vad, chen2024vadv2, transfuser, roach} or to function as standalone decision-makers~\cite{drivegpt4, driving-with-llms, languagempc, hwang2024emma, tian2024drivevlm}, as they are assumed to be able to transfer the generalization ability to the road scenes \cite{achiam2023gpt4,wang2024drivecot,shao2024lmdrive,sima2024drivelm,wang2024he,nie2024reason2drive,shao2023reasonnet}. Furthermore, novel approaches leverage collaborative LLM-agents for editable scene simulation, offering new paradigms for data generation. \cite{wei2024editable} In this line of research, some approaches translate structured driving inputs—such as perception outputs and HD maps—into language for planning~\cite{driving-with-llms, languagempc}, while others like DriveGPT4~\cite{drivegpt4} process front-camera video to predict both control commands and rationales. LVLM-based planners have also been validated in simulation environments such as CARLA~\cite{dosovitskiy2017carla, drivemlm}, and large-scale pretraining (e.g., ELM~\cite{zhou2024elm}) has shown promise for improving generalization. Recent works further propose driving-specific Q\&A data and benchmarks~\cite{sima2023drivelm, wu2023nuprompt, qian2024nuscenes-qa, wu2023refer-kitti, deruyttere2019talk2car, kim2018bdd-x} to better align training with downstream planning tasks.

\textbf{Specialized Techniques and Datasets for Autonomous Driving.} Beyond the VLM paradigm, significant research continues to advance various critical aspects of autonomous driving systems, addressing specific challenges in perception, simulation, mapping, and prediction. For instance, in the realm of realistic simulation, Mars~\cite{wu2023mars} offers an instance-aware, modular, and realistic simulator leveraging neural radiance fields, crucial for generating and testing complex scenarios. Challenger~\cite{xu2025challenger} focuses on generating physically plausible yet realistic adversarial driving videos to stress-test AD systems against aggressive maneuvers, while AVD2~\cite{li2025avd2} introduces a novel framework for generating accident videos aligned with detailed natural language descriptions and preventative measures, thereby enhancing accident scenario understanding for training and analysis. To enhance robustness in challenging environmental conditions, especially at night, joint self-supervised nighttime image enhancement and depth estimation are explored by Steps~\cite{zheng2023steps}, crucial for improved visual perception in low-light settings. Accurate and far-reaching environmental representation is further enabled by P-MapNet~\cite{jiang2024p}, which leverages both standard definition (SDMap) and high-definition (HDMap) priors for superior map generation, improving situational awareness over longer distances. In the domain of 3D scene understanding from limited inputs, MonoOcc~\cite{zheng2024monoocc} delves into monocular semantic occupancy prediction, aiming to reconstruct comprehensive 3D geometry and semantics from single-camera views. For robust motion forecasting, especially in dynamic multi-agent environments, Int2~\cite{yan2023int2} presents a large-scale dataset and framework specifically for interactive trajectory prediction at complex intersections, capturing crucial dynamics that are vital for safe navigation. These targeted innovations collectively pave the way for more capable and reliable autonomous driving systems. Efforts also extend to generating high-quality, annotated training data, with UniScene~\cite{li2024uniscene} proposing a unified occupancy-centric framework for comprehensive driving scene generation. Furthermore, SCP-Diff~\cite{gao2024scp} significantly improves the quality of semantic image synthesis for sensor simulation by introducing a spatial-categorical joint prior, enabling the creation of highly realistic and diverse virtual environments with precise semantic control. These targeted innovations collectively pave the way for more capable and reliable autonomous driving systems.

\textbf{End-to-end Autonomous Driving Datasets and Benchmarks.}
We categorize autonomous driving benchmarks into two categories, one for large-scale imitation learning, and one for simulation. The first category includes large-scale, real-world datasets, often collected from road networks \cite{caesar2020nuscenes,nuplan,Once,neuhold2017mapillary,sun2020scalability,varma2019idd,argoversev2}, which are crucial for developing and evaluating systems on annotated perception, prediction, and planning tasks. 
In this work, we select representative imitation learning benchmarks for constructing our dataset:
KITTI \cite{kitti}, an early benchmark, provided data from Germany. nuScenes \cite{caesar2020nuscenes} expanded on this with data from Boston and Singapore. The Waymo Open Dataset \cite{sun2020scalability} offers immense scale with data collected from diverse US locations. Argoverse (v1 \& v2) \cite{argoversev1,argoversev2} also features data from various US cities. nuPlan provides over 1200 hours of driving data from cities in the US and Singapore. For global visual diversity, Mapillary Vistas \cite{neuhold2017mapillary} includes street-level imagery from all continents. ONCE \cite{Once} contributes a massive dataset with 1 million LiDAR scenes and 7 million camera images from China. Finally, the India Driving Dataset provides crucial data from challenging and unstructured driving environments across India.\cite{varma2019idd,dokania2023idd}
The second line involves simulation-based benchmarks, such as Bench2Drive \cite{jia2024bench2drive}, NAVSIM \cite{dauner2024navsim}, and NeuroNCAP \cite{ljungbergh2024neuroncap}, which offer closed-loop evaluation environments. These simulators utilize metrics more akin to driving task-oriented reward designs, allowing for systematic testing of decision-making and control algorithms in interactive scenarios.
Notably, our dataset construction prioritizes the collection and filtering of authentic, real-world unstructured scenarios, rather than introducing synthetic elements or anomalies \cite{tian2023unsupervised,bogdoll2024umad,tian2024latency,di2021pixel,lis2019detecting,blum2019fishyscapes,pinggera2016lost,xia2020synthesize},. This commitment to genuine data ensures the Impromptu VLA Dataset fosters the development of VLA models grounded in the true complexities of diverse driving conditions.

\section{Conclusion}

This paper introduced the Impromptu VLA Dataset, a meticulously curated benchmark of approximately 80,000 clips featuring rich multi-task Question-Answering annotations and corresponding action trajectories, specifically designed to address the critical data scarcity for autonomous driving in unstructured environments. 
Our comprehensive experiments demonstrate that Vision-Language Models trained with the Impromptu VLA Dataset achieve significant performance gains, evidenced by enhanced closed-loop safety and driving scores on the NeuroNCAP benchmark, as well as improved open-loop trajectory prediction accuracy on nuScenes. Furthermore, evaluations on our dataset's validation suite confirm its efficacy as a diagnostic tool, revealing specific model advancements in perception, prediction, and planning capabilities when handling diverse and challenging unstructured road scenarios. The Impromptu VLA Dataset thus offers a valuable new resource to foster the development of more robust, adaptable, and capable autonomous driving systems prepared for the complexities of real-world operation.
\textbf{Limitation.} We acknowledge that the primary reliance on Qwen2.5-VL for annotation generation in the Impromptu VLA Dataset might introduce potential model-specific biases; however, we believe the holistic human verification and the demonstrated utility in enhancing Vision-Language Model performance in unstructured scenarios confirms its significant value as a research resource.

\newpage

\newpage
\appendix 

\setcounter{subsection}{0}
\section*{Supplementary Material}

This supplementary document provides additional insights and technical details regarding our proposed Impromptu VLA dataset and its associated models. We will begin by detailing the implementation specifics of our training process, including data formats for different experimental stages and hyperparameter settings used for the Qwen2.5-VL model variants. Subsequently, we will elaborate on our construction for the dataset of unstructured driving scenarios, outlining the three key phases: Data exploration, Unconventional Driving Scenario Identification, and Iterative refinement. Finally, we will offer concrete Question-Answering (Q\&A) examples from the Impromptu VLA dataset to illustrate its structure, highlighting key considerations and our approach to end-to-end trajectory prediction.

\section{Implementation Details}
In this section, we provide further details regarding the training processes, including the data format specifically employed when fine-tuning on the nuScenes dataset, and the general hyperparameter settings used for the Qwen2.5-VL model variants during our experiments.

\subsection{Data format for nuScenes Fine-tuning and Evaluation}
\begin{figure}
    \centering
    \includegraphics[width=1.0\linewidth]{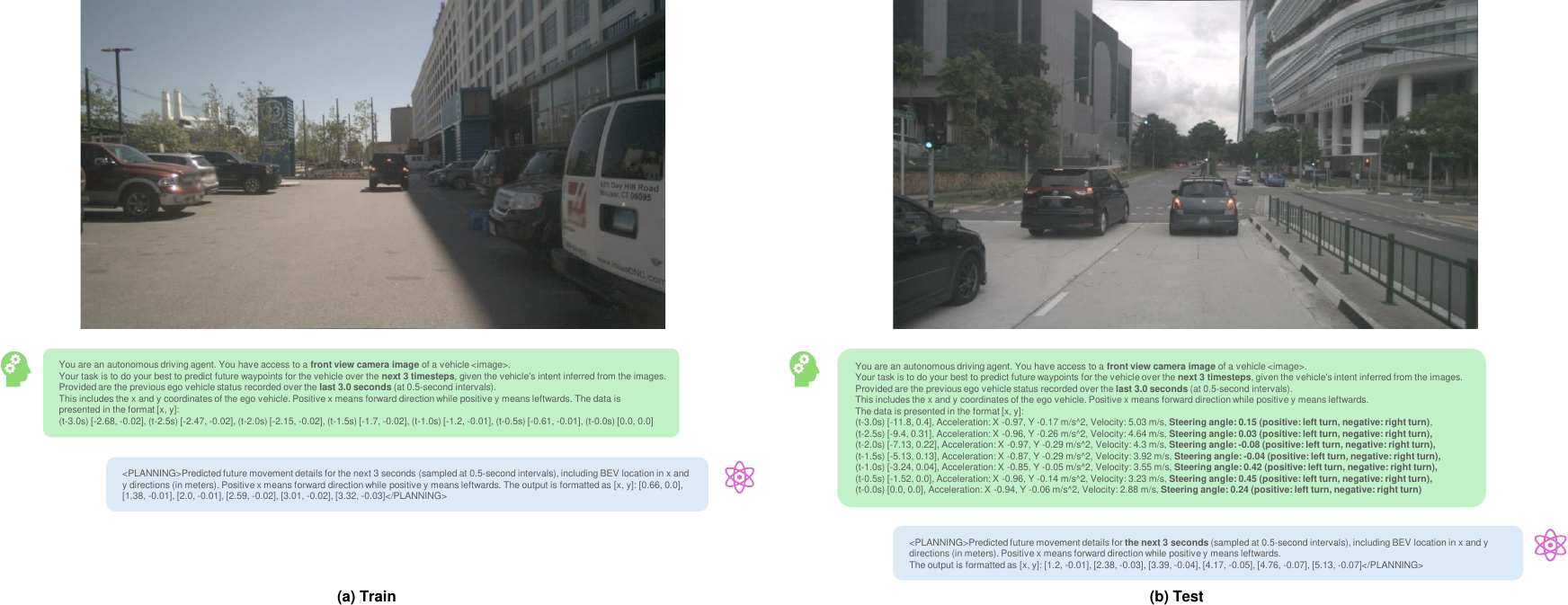}
    \caption{Example of data used when training and testing on nuScenes}
    \label{fig:finetune-example}
\end{figure}

The data format described here pertains specifically to the fine-tuning and evaluation stages performed directly on the nuScenes dataset, as referenced in our main paper's open-loop and closed-loop experiments. 
While the trajectory data format shares similarities with the End-to-End Trajectory Prediction Q\&A format within our Impromptu VLA dataset (which uses past 1.5s and future 5s trajectories), there are distinctions for this nuScenes-specific setup. These include utilizing only the front-camera view and using trajectories spanning the past 3 seconds and future 3 seconds. For this nuScenes-specific training, input consists of the ego-vehicle's past trajectory points. For testing on nuScenes trajectory prediction benchmarks, in addition to past trajectory, velocity, and acceleration (similar to our Impromptu VLA Q\&A format), we also incorporate steering wheel angle information where available from the nuScenes dataset. An example of the data format used for nuScenes experiments is illustrated in Figure \ref{fig:finetune-example}.

\subsection{Hyperparameters}
Our empirical hyperparameter values, outlined in Table \ref{tab:hyperparameters}, were established for training models using the Impromptu VLA Dataset and subsequently on nuScenes data. These values stem from general observation and common practices, not a thorough optimization process, due to the extensive search space. Nevertheless, with these settings, our models demonstrate a superior grasp of unstructured environments, leading to a significant improvement in trajectory prediction, as shown in the main paper. We recognize that further refinement through exhaustive tuning could yield even better results.

\begin{table}[h!]
\centering
\begin{tabular}{ll}
    \toprule 
    Hyperparameter & Value \\
    \midrule 
    Cutoff len & 4096 \\
    Finetuning type & full \\
    Image resolution & 262144 \\
    Learning rate & 5.0e-06 \\
    Scheduler type & cosine \\
    Warmup ratio & 0.03 \\
    \bottomrule 
    \end{tabular}
    \caption{\textbf{Hyperparameters used in training}. This table details the empirical values for crucial hyperparameters employed across our entire pipeline. These settings were consistently applied in all experiments and were derived from general observations rather than specific task-based tuning.}
    \label{tab:hyperparameters}
\end{table}

\section{VLM-based Unstructured Scene Identification}
Following the Data Exploration phase where rich textual descriptions were generated for each scene, our objective was to isolate the scenarios that genuinely represented "unstructured" or "unconventional" driving conditions. This was critical for focusing the dataset on true corner cases. To achieve this, we employed a VLM-based classification step. Each scene description generated by Qwen2.5-VL was evaluated using a carefully designed prompt, which instructed the VLM to act as a scenario categorizer and determine if the description corresponded to an unconventional case (as opposed to a routine, structured driving situation). 

\begin{figure}
    \centering
    \includegraphics[width=1.0\linewidth]{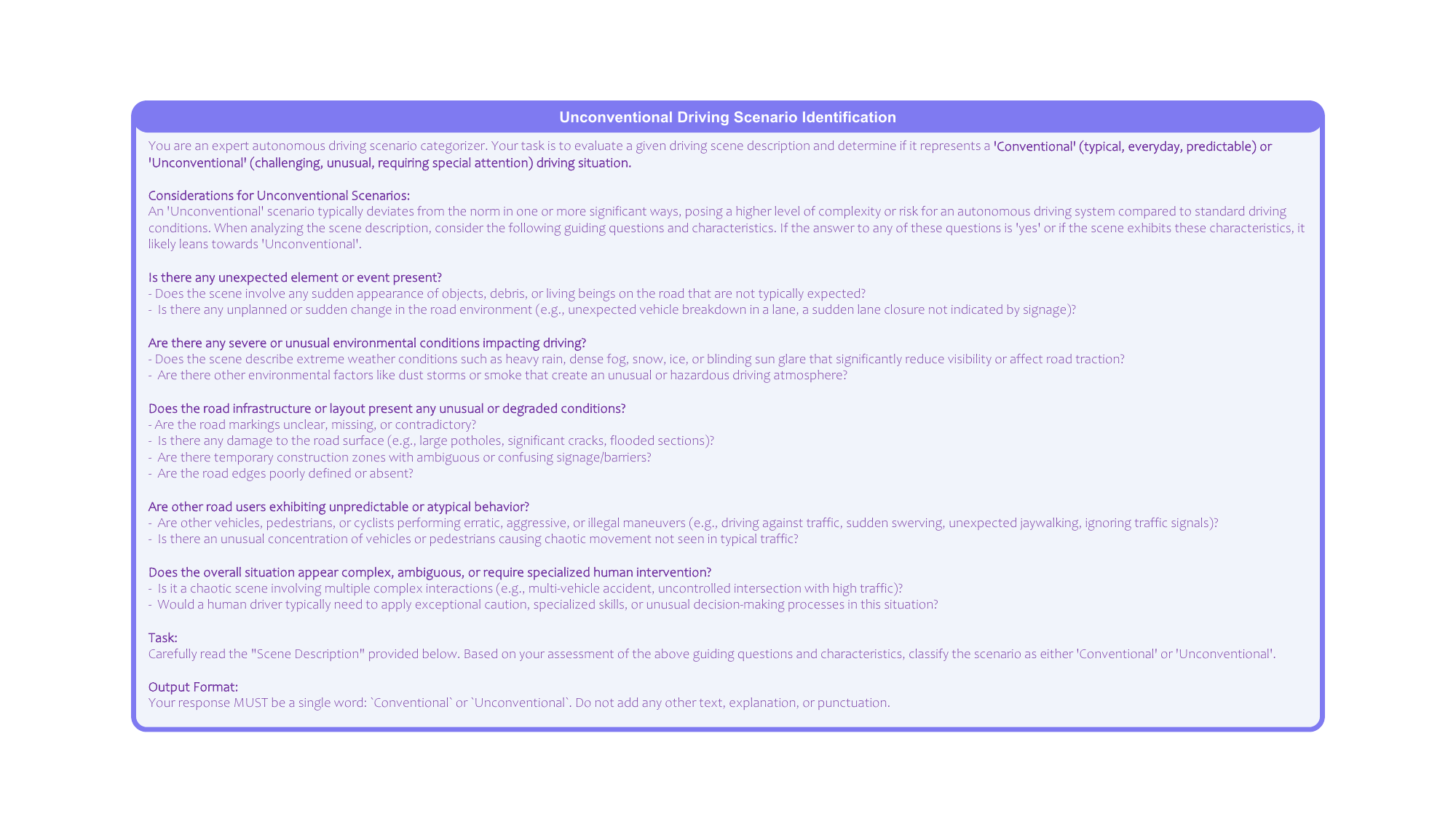}
    \caption{Prompt used during Unconventional Driving Scenario Identification}
    \label{fig:Identification}
\end{figure}

The effectiveness of this VLM-based filtering prompt was crucial. Therefore, we conducted an iterative refinement process on a validation subset of approximately 1000 scene descriptions, which were also independently labeled by two human annotators as 'Conventional' or 'Unconventional'. The VLM's classifications were compared against this human consensus, and the prompt (an example of which is shown in Figure \ref{fig:Identification}) was iteratively adjusted until a high degree of agreement was achieved. This ensured that the scenarios passed to the next stage were indeed representative of the complex conditions we aimed to capture.

\section{Q\&A Examples}
\begin{figure}
    \centering
    \makebox[\linewidth]{%
        \includegraphics[width=1.2\linewidth]{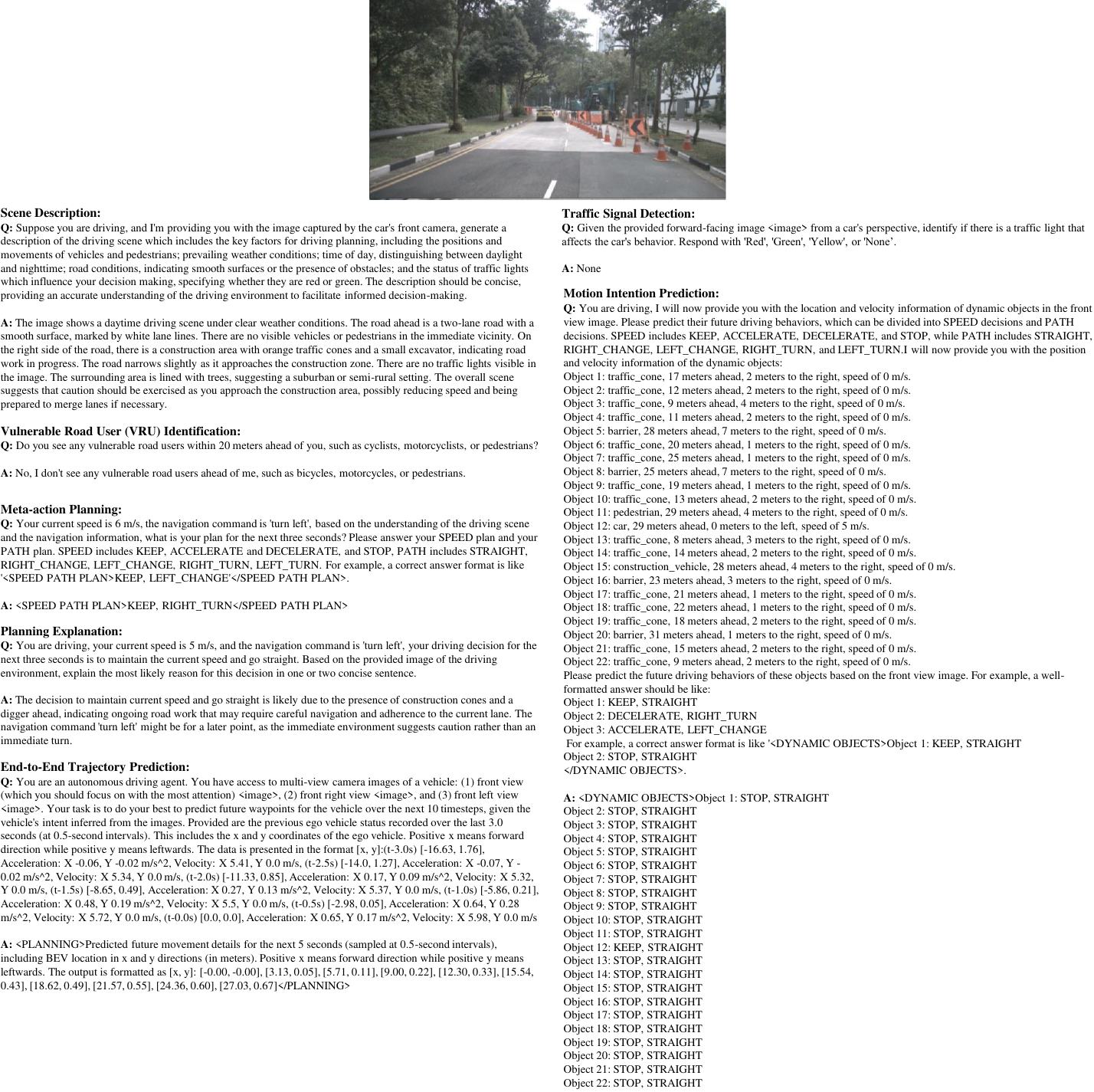}%
    }
    \caption{Q\&A example from the nuScenes dataset}
    \label{fig:QA1}
\end{figure}

\begin{figure}
    \centering
    \makebox[\linewidth]{%
        \includegraphics[width=1.2\linewidth]{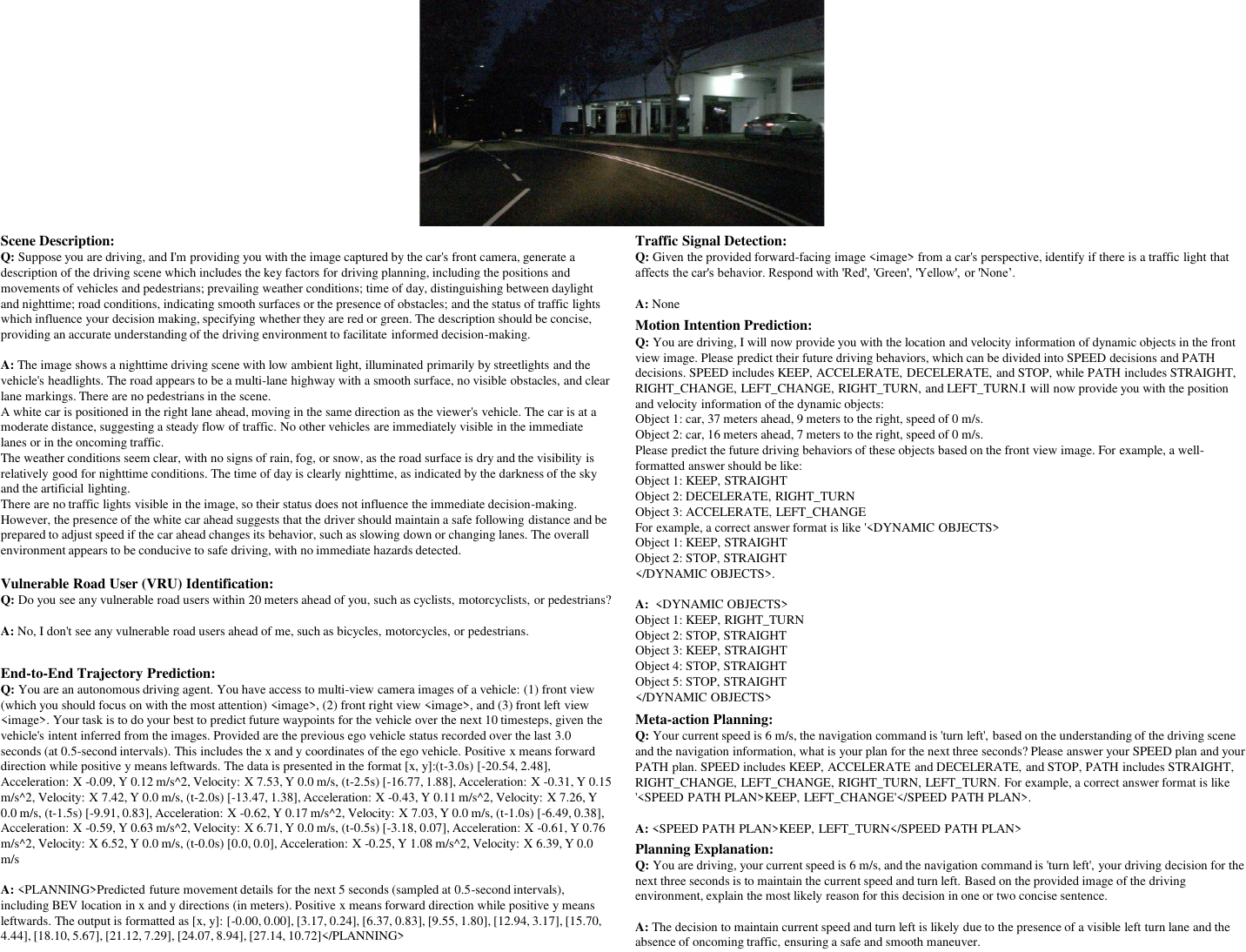}%
    }
    \caption{Q\&A example from the nuScenes dataset}
    \label{fig:QA2}
\end{figure}

\begin{figure}
    \centering
    \makebox[\linewidth]{%
        \includegraphics[width=1.2\linewidth]{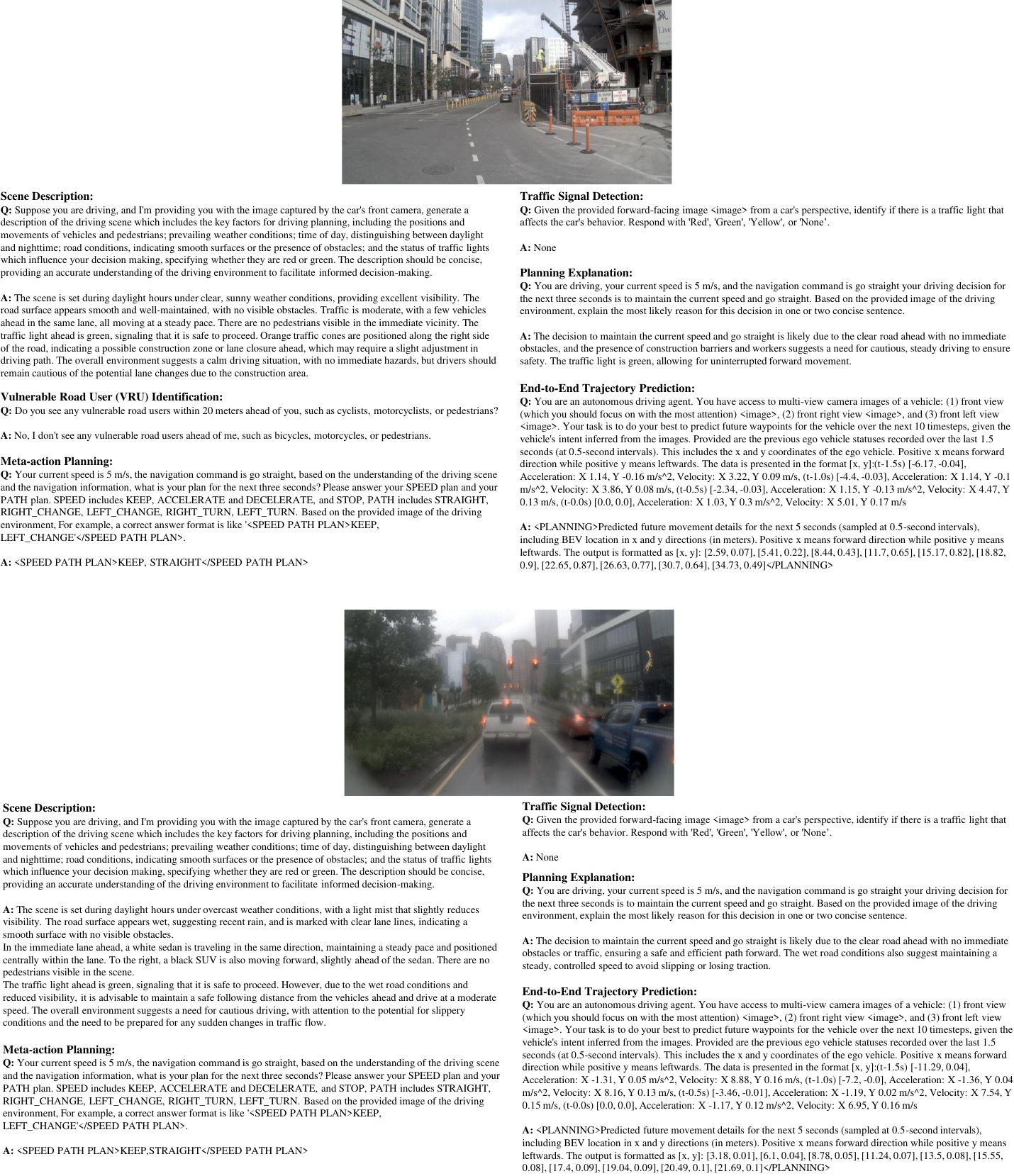}%
    }
    \caption{Q\&A examples from the NAVSIM dataset}
    \label{fig:QA3}
\end{figure}

\begin{figure}
    \centering
    \makebox[\linewidth]{%
        \includegraphics[width=1.2\linewidth]{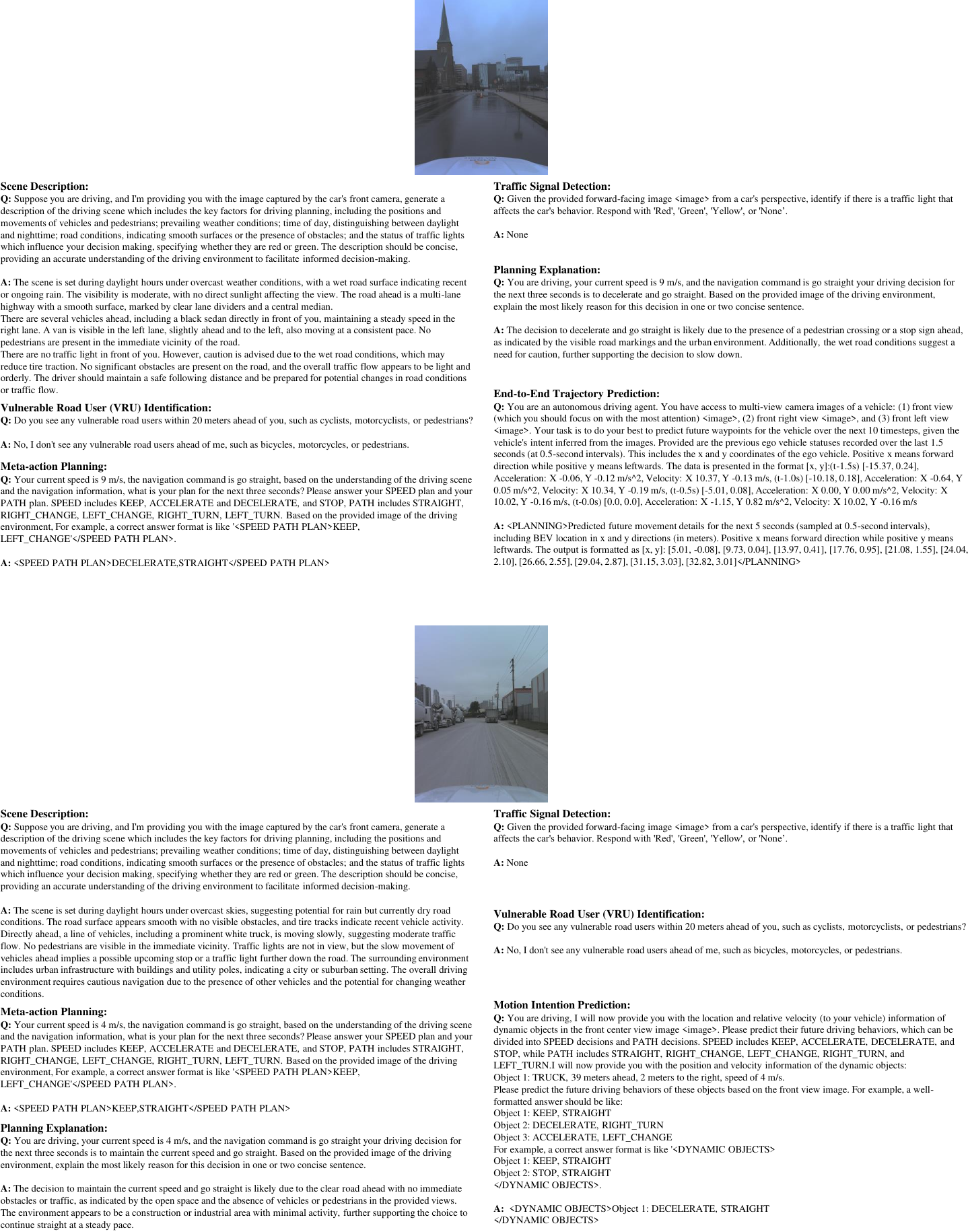}%
    }
    \caption{Q\&A examples from the Argoverse-V2 dataset}
    \label{fig:QA4}
\end{figure}

\begin{figure}
    \centering
    \makebox[\linewidth]{%
        \includegraphics[width=1.2\linewidth]{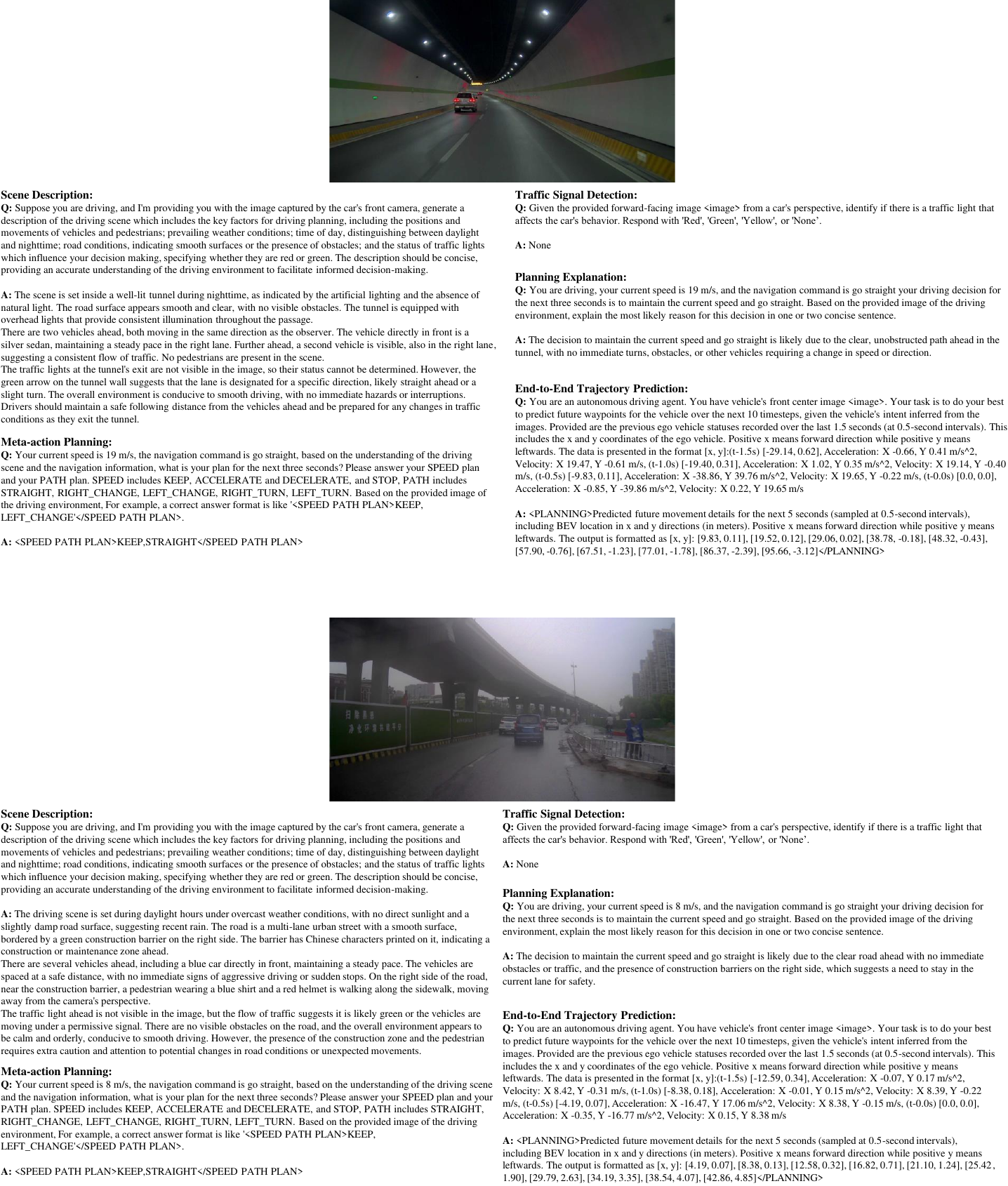}%
    }
    \caption{Q\&A examples from the ONCE dataset}
    \label{fig:QA5}
\end{figure}

\begin{figure}
    \centering
    \makebox[\linewidth]{%
        \includegraphics[width=1.2\linewidth]{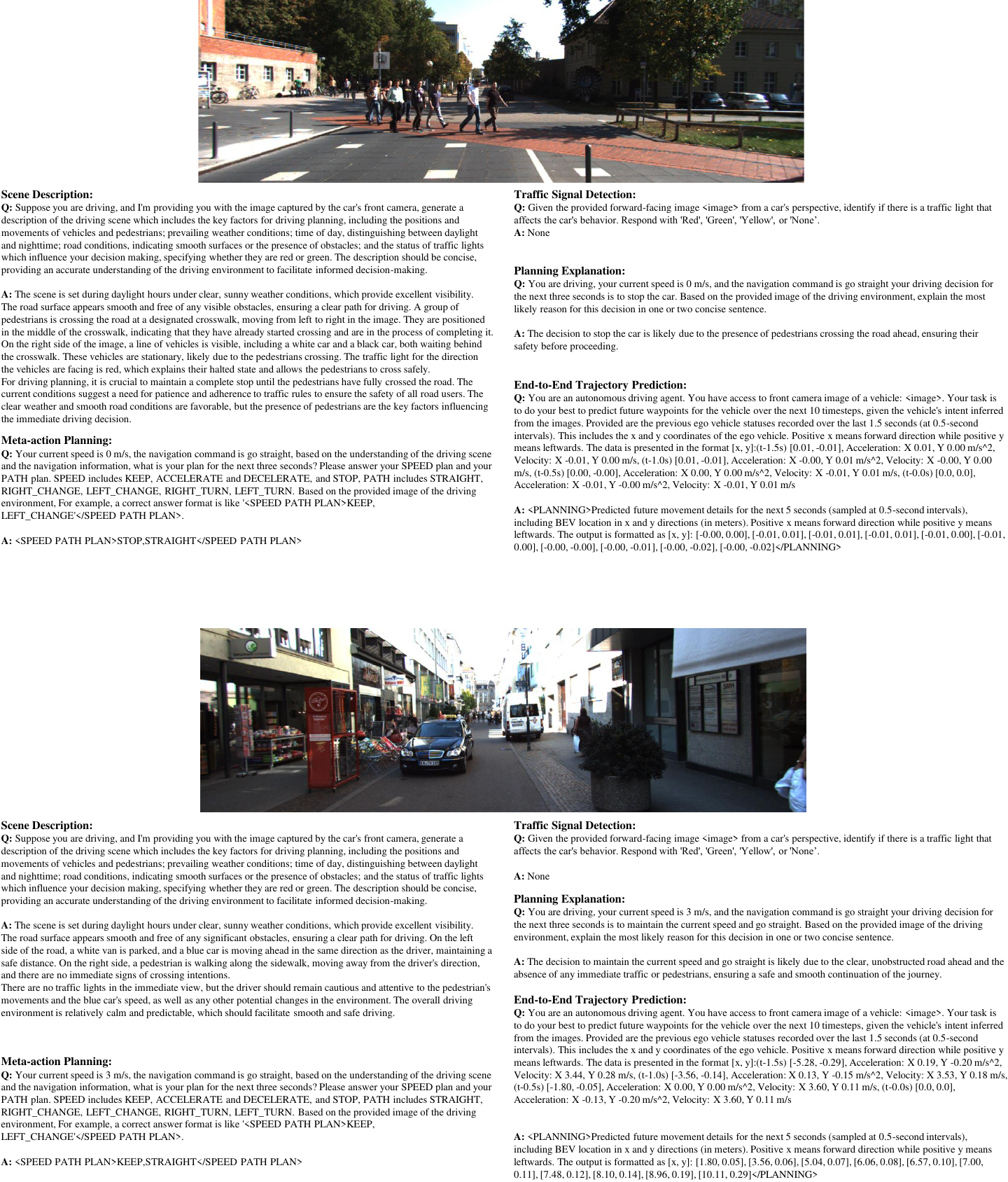}%
    }
    \caption{Q\&A examples from the KITTI dataset}
    \label{fig:QA6}
\end{figure}

\begin{figure}
    \centering
    \makebox[\linewidth]{%
        \includegraphics[width=1.2\linewidth]{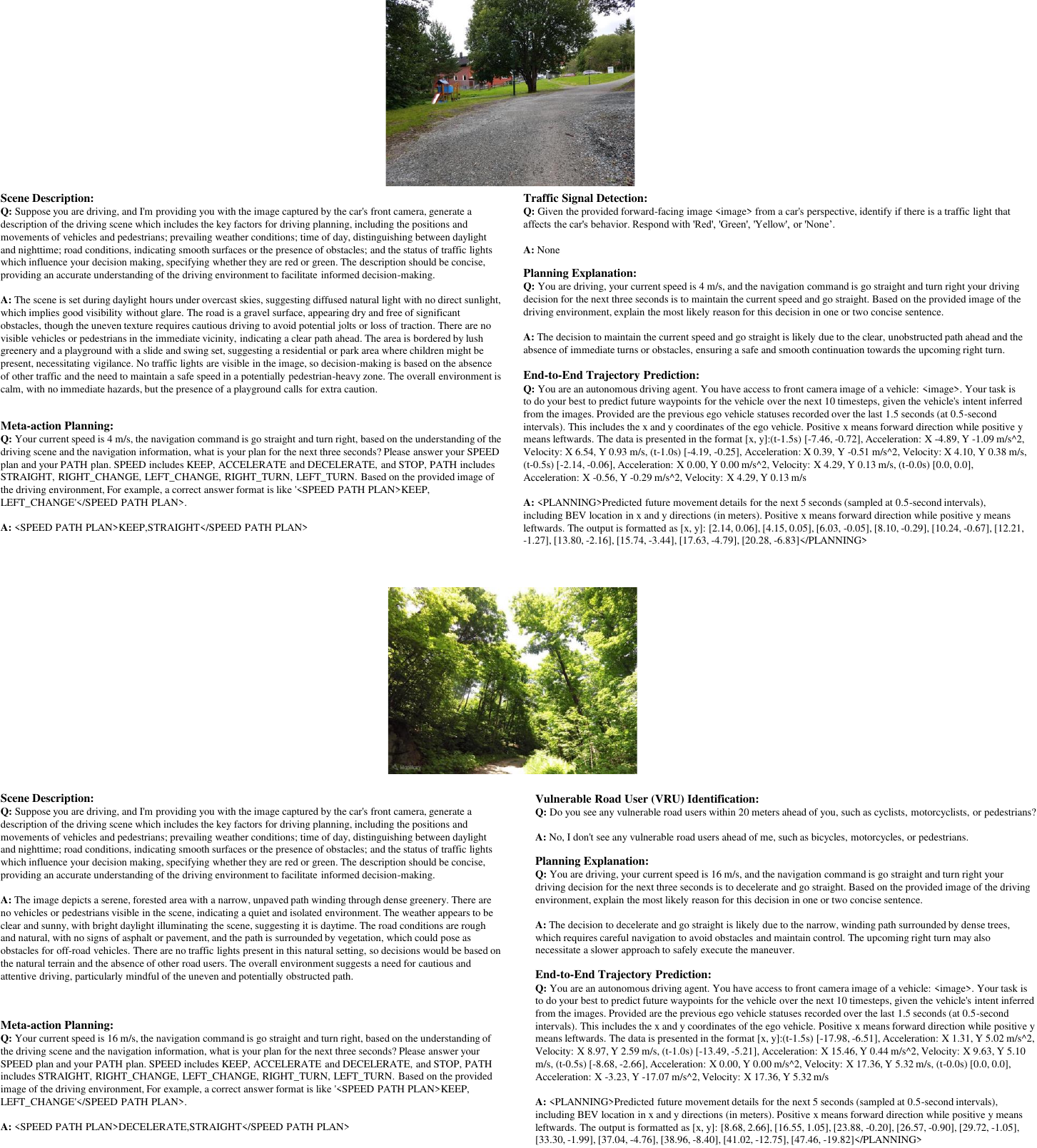}%
    }
    \caption{Q\&A examples from the Mapillary dataset}
    \label{fig:QA7}
\end{figure}

\begin{figure}
    \centering
    \makebox[\linewidth]{%
        \includegraphics[width=1.2\linewidth]{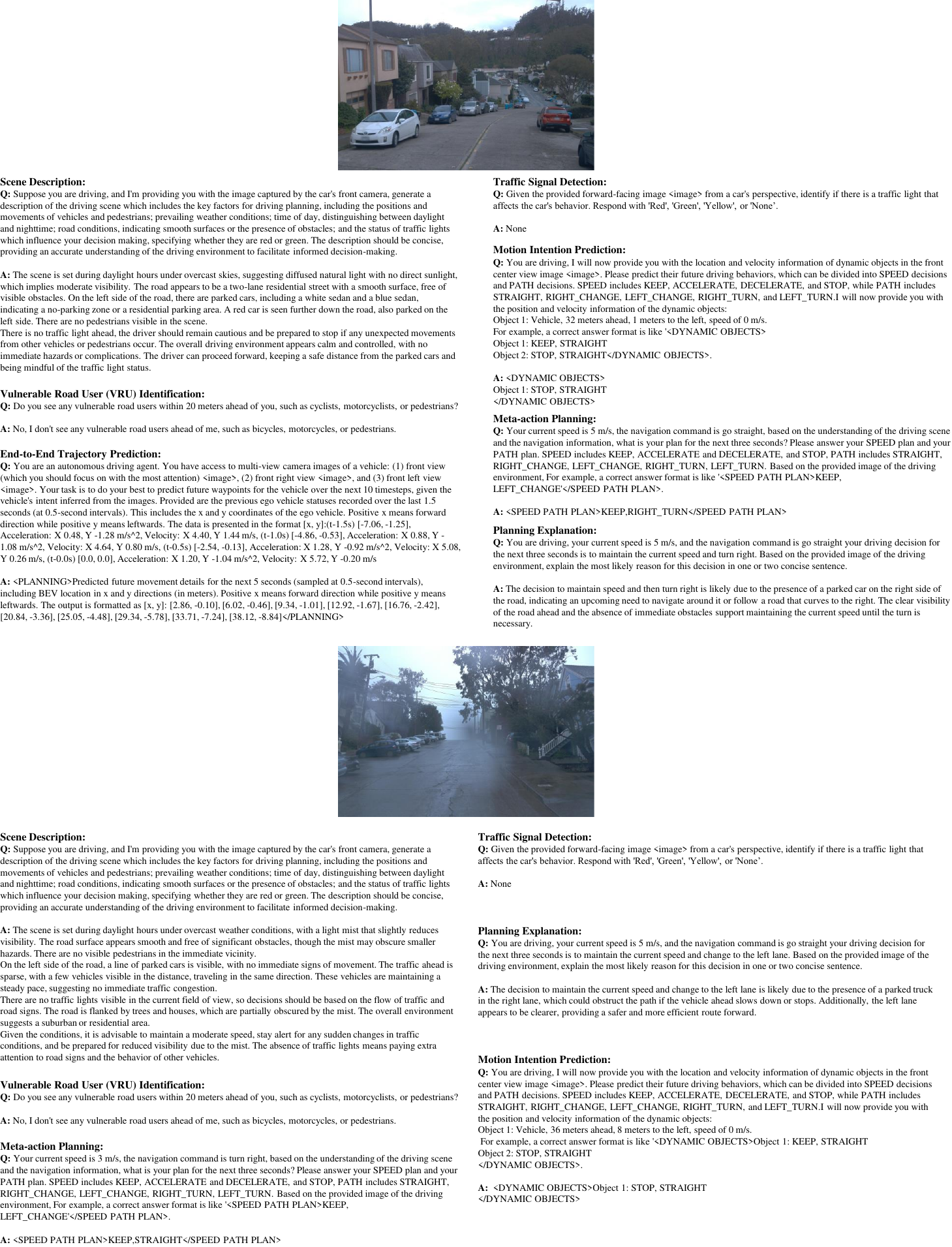}%
    }
    \caption{Q\&A examples from the Waymo dataset}
    \label{fig:QA8}
\end{figure}

\begin{figure}
    \centering
    \makebox[\linewidth]{%
        \includegraphics[width=1.2\linewidth]{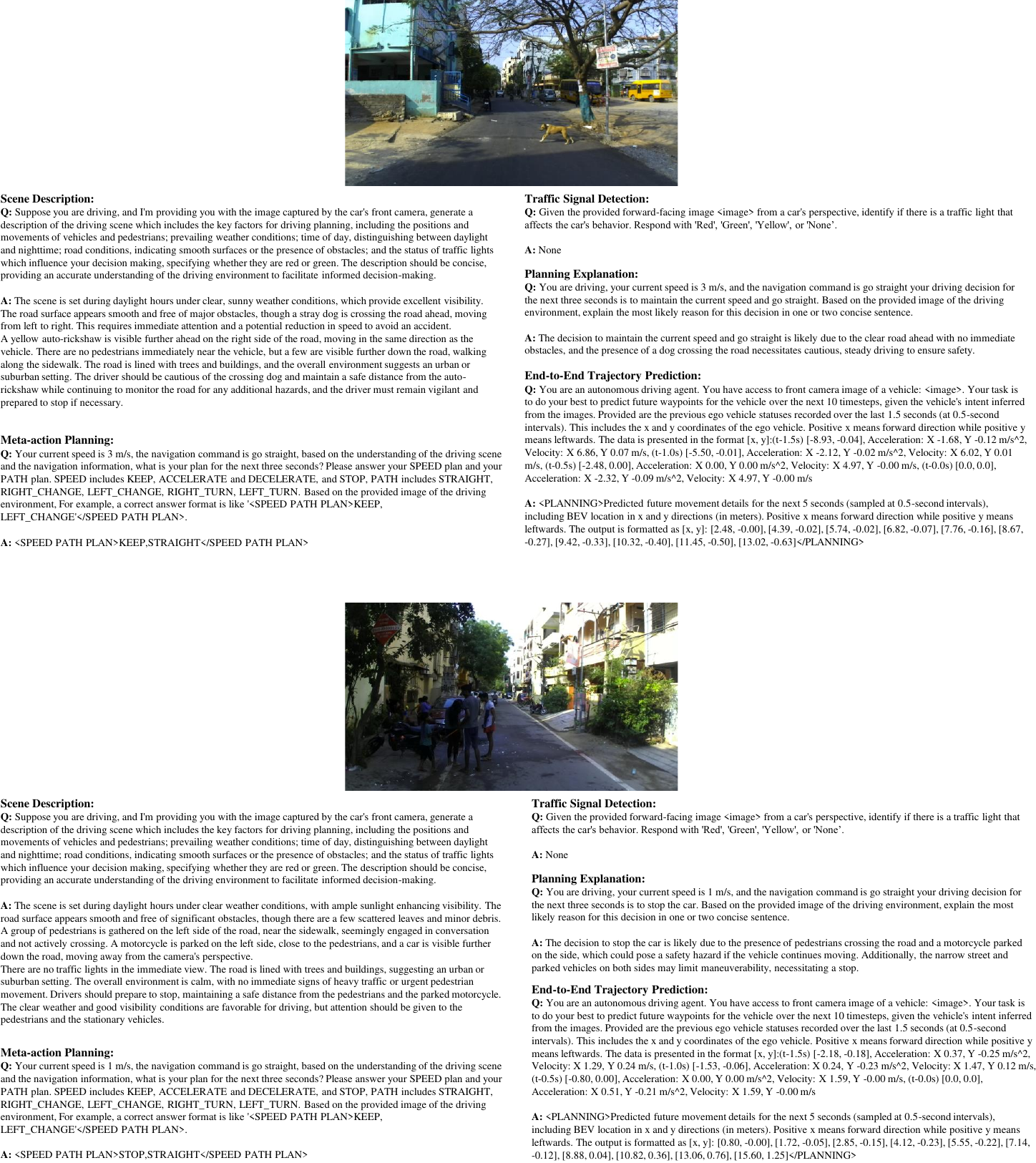}%
    }
    \caption{Q\&A examples from the IDD dataset}
    \label{fig:QA9}
\end{figure}

Our Question-Answering (QA) format for the Impromptu VLA dataset, inspired by frameworks like Senna, is designed to be comprehensive. We present two distinct QA data examples from various source datasets that contribute to Impromptu VLA, as illustrated in Figures \ref{fig:QA1} through \ref{fig:QA9}. These examples include the corresponding front-facing images and detailed QA information.

A notable aspect of our QA format is the inclusion of special tokens to differentiate between tasks with similar output structures. For instance, tokens like \texttt{<PLANNING>} and \texttt{<DYNAMIC\_OBJECTS>} are used. We found through experimentation that without such disambiguation, the model could confuse, for example, the ego-vehicle's meta-action plan with the predicted motion intentions for other dynamic objects in the scene, due to the structural similarity of their speed and path plan outputs. These special tokens ensure the model can distinguish the context effectively.

The Impromptu VLA dataset includes an end-to-end trajectory prediction Q\&A task. As detailed in the main paper, this incorporates the ego-vehicle's past 1.5 seconds of states (including displacement, velocity, and acceleration relative to its current position) and its ground-truth future 5-second trajectory. This formulation is designed to facilitate the model's learning of long-range trajectory prediction, a key capability that also supports tasks like understanding and executing navigation-related commands.

Additionally, it is important to note that the availability of all defined Q\&A tasks for a given clip in the Impromptu VLA dataset depends on the richness of the annotations in the source dataset from which the clip was curated. For instance, if a source clip lacks ground-truth object detection data, generating Q\&A pairs for tasks like Vulnerable Road User (V.R.U.) Identification (which relies on ground truth [cite: 82]) or detailed Motion Intention Prediction for specific actors may not be possible for that particular clip, even if other Q\&A tasks like scene description or ego-trajectory prediction can be generated.

\end{document}